\documentclass[11pt,letterpaper]{article}
\usepackage[T1]{fontenc}
\usepackage[utf8]{inputenc}
\usepackage{mathptmx}
\usepackage[scaled=0.92]{helvet}
\usepackage[margin=1in]{geometry}
\usepackage{microtype}
\usepackage{graphicx}
\usepackage{booktabs}
\usepackage{tabularx}
\usepackage{array}
\usepackage{ragged2e}
\usepackage{enumitem}
\usepackage{caption}
\usepackage{xurl}
\usepackage[hidelinks,hyperfootnotes=false]{hyperref}

\hypersetup{
  pdftitle={Practitioner Beliefs and Behaviors in AI-Enhanced Education: DOT Framework Survey Evidence},
  pdfauthor={David Gibson, M. Elizabeth Azukas, Gerald Knezek},
  pdfkeywords={artificial intelligence, higher education, DOT Framework, design thinking, open systems theory, practitioner beliefs, exploratory factor analysis}
}

\newcommand{\floatnote}[1]{\par\vspace{4pt}{\footnotesize\RaggedRight #1\par}}

\newcommand{\unmarkedfootnote}[1]{\begingroup\renewcommand{\thefootnote}{}\footnotetext{#1}\endgroup}

\begin{document}

\begin{center}
{\LARGE\bfseries Practitioner Beliefs and Behaviors in AI-Enhanced Education: DOT Framework Survey Evidence\par}
\vspace{1.4em}
\begin{tabular}[t]{c}
{\large David Gibson}\\
Curtin University, Australia\\
\href{mailto:davidcgibson50@gmail.com}{davidcgibson50@gmail.com}
\end{tabular}

\vspace{0.9em}
\begin{tabular}[t]{c}
{\large M. Elizabeth Azukas}\\
Georgia Tech Research Institute, Georgia Institute of Technology, USA\\
\href{mailto:elizabeth.azukas@gtri.gatech.edu}{elizabeth.azukas@gtri.gatech.edu}
\end{tabular}

\vspace{0.9em}
\begin{tabular}[t]{c}
{\large Gerald Knezek}\\
University of North Texas, USA\\
\href{mailto:gknezek@gmail.com}{gknezek@gmail.com}
\end{tabular}
\end{center}

\unmarkedfootnote{\textbf{Author Note.} M. Elizabeth Azukas conducted this research while affiliated with East Stroudsburg University, USA; she is now at Georgia Tech Research Institute, Georgia Institute of Technology. The study was approved by the Institutional Review Board of East Stroudsburg University. An earlier version of this manuscript was posted as a preprint on arXiv (\url{https://arxiv.org/abs/2605.29041}); the preprint will be updated to reflect this revision. Correspondence concerning this article should be addressed to David Gibson, Curtin University. Email: davidcgibson50@gmail.com}

\vspace{0.6em}
\begin{abstract}
\noindent Generative artificial intelligence (AI) is spreading through higher education faster than the institutional structures that guide its use, and little is known about how practitioners' beliefs and practices align with design-oriented frameworks for AI integration. We report an exploratory cross-sectional survey of 72 higher education practitioners, interpreted through the DOT Framework, which combines design thinking with open systems theory. We examined AI familiarity and use, design-oriented practices, the structure of beliefs about AI's pedagogical role, institutional conditions, and differences by AI familiarity. Respondents used AI mainly to generate instructional materials and support research, and rarely for grading or assessment. Experimenting with prompts was common, whereas collecting feedback and iterating on AI outputs was the least frequent design practice. Exploratory factor analysis of 19 belief items yielded three factors: AI Functional Capabilities, Oversight and Governance, and Instructor Collaboration and Planning. Oversight beliefs were near-universal and did not vary with familiarity, whereas familiarity was associated with use, design practices, and capability beliefs. Ethical concerns, training, and policy gaps were the most frequently reported institutional challenges. The findings are consistent with, but do not confirm, the framework's distinctions and indicate where confirmatory and outcome-focused research is needed.

\medskip
\noindent \textit{Keywords:} artificial intelligence, higher education, DOT Framework, design thinking, open systems theory, practitioner beliefs, exploratory factor analysis
\end{abstract}

\vspace{0.6em}
Generative artificial intelligence (AI) tools have entered higher education teaching, research, and administration within a few years, and instructors now make daily decisions about whether and how to use them. Institutions have been slower to respond. Reviews of the literature identify assessment integrity, ethical considerations, and policy development as areas needing further work (Batista et al., 2024) and call for greater ethical, methodological, and contextual rigor in research on AI in higher education (Bond et al., 2024). Interview research across multiple universities reports that instructors need more education about the tools available to them (Baytas \& Ruediger, 2025). In this situation, practitioners often build their own practices with limited guidance, and those practices are likely to vary widely in intentionality and alignment with educational goals.

We regard this as a problem of practice rather than a purely technical one. Without shared structures for designing, testing, and evaluating AI-supported instruction, individual experimentation is difficult to learn from, to scale, or to hold accountable. Conceptual frameworks have been proposed to address this problem. One of them, the DOT Framework (Azukas \& Gibson, 2025), combines Design Thinking (D) with the Open Systems model (O) to guide the Transformation (T) of teaching through intentional, human-centered AI use at both classroom and institutional levels. Like most frameworks in this area, however, it was developed conceptually, and there is little evidence about whether practitioners' beliefs and reported practices correspond to the distinctions it draws.

The purpose of this study is to describe the AI-related beliefs, practices, and institutional conditions reported by higher education practitioners and to examine how these correspond to the constructs of the DOT Framework. Five research questions guided the study:

\begin{description}[leftmargin=3.2em, labelwidth=2.6em, labelsep=0.6em, font=\normalfont, itemsep=2pt, topsep=4pt]

\item[RQ1.] How familiar are higher education practitioners with AI, and how do they currently use it across instructional, assessment, research, and administrative tasks?

\item[RQ2.] To what extent do practitioners report design-oriented practices (needs identification, goal specification, experimentation, and feedback and iteration) and verification-oriented prompting techniques when using AI?

\item[RQ3.] What latent structure underlies practitioners' beliefs about AI's pedagogical role, and how does that structure correspond to the constructs of the DOT Framework?

\item[RQ4.] What institutional challenges, concerns, and support needs do practitioners report?

\item[RQ5.] Do beliefs, AI use, and design-oriented practices differ by level of AI familiarity?

\end{description}

The study makes three contributions. It provides the first empirical description of practitioner beliefs and practices organized around the DOT Framework; it reports an exploratory measurement structure for beliefs about AI's pedagogical role; and it identifies where reported practice diverges from the full design cycle that the framework recommends. All three contributions are exploratory and are intended to inform confirmatory and outcome-focused research.

\section{Literature Review}

\subsection{Key Constructs}

Four constructs guide this study. \textit{Practitioner beliefs about AI} are the propositions that educators hold to be true about what AI can and should do in teaching. We draw on the teacher-beliefs tradition, which argues that pedagogical beliefs help explain why technology use often remains limited even when access, training, and supportive policy are in place (Ertmer, 2005). This framing differs from technology acceptance models, which center on perceived usefulness and ease of use as predictors of intention to use a technology (Davis, 1989). \textit{AI use} refers to the tasks for which practitioners report using AI and how often. \textit{Design-oriented practice} refers to the extent to which AI use follows a deliberate cycle of understanding needs, defining goals, generating and testing options, and revising in light of feedback; design thinking offers teachers an accessible structure for approaching problems of practice in this way (Henriksen et al., 2017). \textit{Institutional conditions} are the policies, training, infrastructure, and norms that enable or constrain classroom-level practice, a concern central to socio-technical systems theory (Mumford, 2000; Trist \& Bamforth, 1951).

\subsection{Research on Faculty Use of and Beliefs About AI}

Survey research on faculty and generative AI has grown quickly. Mah and Groß (2024) surveyed 122 faculty members and identified four latent profiles (optimistic, critical, critically reflected, and neutral); respondents viewed greater equity as AI's greatest benefit, named limited AI literacy among the greatest challenges, and most expressed interest in professional development. Lyu et al. (2025) used mixed methods with 178 instructors at one U.S. university to examine practices, perceptions, and trust and distrust of generative AI. Nevárez Montes and Elizondo-Garcia (2025) applied the technology acceptance model to 208 faculty members at a private university in Mexico, and Almisad and Aleidan (2025) reported that 102 faculty members in Kuwait perceived benefits in reducing administrative tasks, supporting research, and innovating curriculum design. Earlier structural equation modeling by Wang et al. (2021) found that teachers' self-efficacy, attitudes, perceived ease of use, perceived usefulness, and anxiety predicted their intention to learn to use AI-based applications. Interview research conducted with 19 institutions documented a need for more education about the generative AI product landscape (Baytas \& Ruediger, 2025), and the same research group has catalogued the products marketed to faculty and students (Baytas \& Ruediger, 2024).

\subsection{Agreement, Tension, and Limitations}

Several points of agreement emerge from this work. Faculty see value in AI for content creation, research support, and administrative efficiency, and they raise concerns about ethics and academic integrity and about the adequacy of training and support (Almisad \& Aleidan, 2025; Batista et al., 2024; Mah \& Groß, 2024). Reviews of AI's effects on students' critical thinking similarly conclude that educators need to guide responsible use and the verification of AI outputs (Melisa et al., 2025).

Two tensions are also apparent. First, optimism about AI's benefits coexists with critical and distrustful orientations, sometimes within the same population (Lyu et al., 2025; Mah \& Groß, 2024). Second, much of the quantitative work relies on acceptance constructs that predict whether faculty intend to use AI rather than how they design, test, and evaluate that use (Nevárez Montes \& Elizondo-Garcia, 2025; Wang et al., 2021). The literature also has methodological limitations. Many studies are single-institution and descriptive, instruments are often developed for a single study, and a meta-review of AI in higher education research called for greater ethical, methodological, and contextual rigor (Bond et al., 2024).

\subsection{The Gap Addressed by This Study}

We found no empirical study that examines practitioners' beliefs and practices against a framework that integrates the design process with institutional systems. The DOT Framework was proposed to fill that conceptual role (Azukas \& Gibson, 2025) and has since been extended to leadership contexts (Gibson et al., 2026), but its distinctions have not been examined with practitioner data. This study extends the literature in two ways: it describes design-oriented practice and verification behavior alongside use and beliefs, and it explores whether belief items written to reflect the framework form interpretable dimensions.

\section{Conceptual Framework}

\subsection{The DOT Framework}

The DOT Framework positions design thinking stages against open systems components (Figure 1). Design thinking contributes an iterative sequence of empathizing, defining, ideating, prototyping, and testing (Meinel et al., 2011). Open systems theory, drawing on von Bertalanffy (1950) and Luhmann (1984/1995), contributes the components of environment, input, process, structure, output, and feedback. Each intersection of a stage and a component suggests a design question; for example, Empathize $\times$ Environment grounds design in an understanding of context, and Test $\times$ Feedback supports recursive improvement. The framework applies these intersections at two levels: the micro-system of the classroom, where instructors and students interact with AI tools, and the macro-system of the institution, where policy, infrastructure, and professional development shape what classroom innovation is possible (Azukas \& Gibson, 2025).

\begin{figure}[!htbp]
\caption{The DOT Framework: Design Thinking Stages Positioned Against Open Systems Components}
\label{fig:dot}
\centering
\includegraphics[width=\textwidth]{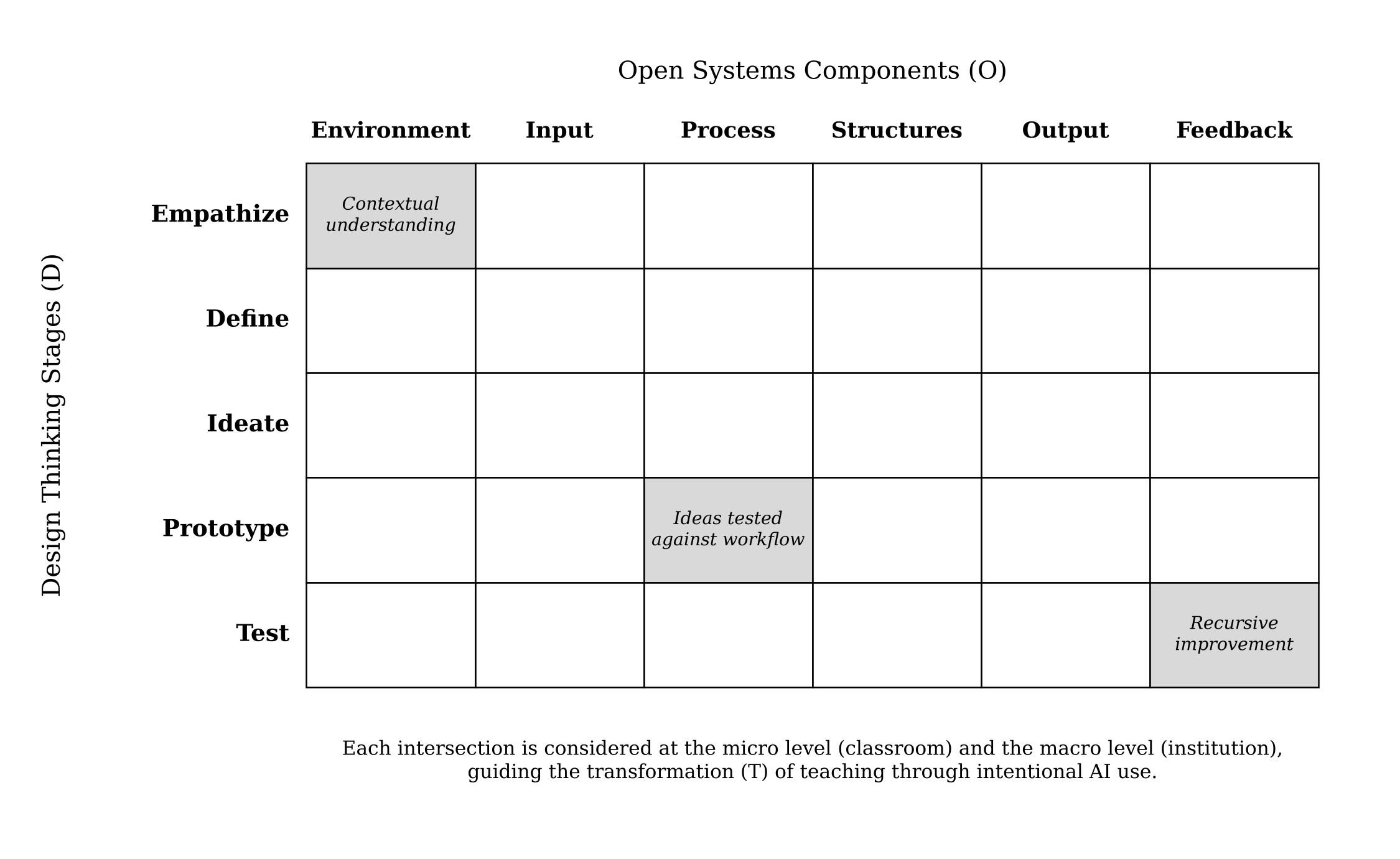}
\floatnote{\textit{Note.} Shaded cells illustrate three intersections discussed in the text. Adapted from Azukas and Gibson (2025).}
\end{figure}

\subsection{AI as a Co-Intelligent Collaborator}

The framework treats AI as a co-intelligent collaborator whose contributions are shaped and evaluated through human guidance (Mollick, 2024). This view extends the idea of distributed intelligence, in which cognition is spread across people and the tools they use (Pea, 1993), to the generative and semi-autonomous properties of large language models. Following Mollick (2024), the framework emphasizes four working principles: treat AI as a capable but fallible collaborator whose work requires review; use AI to support rather than replace higher-order thinking; experiment actively to learn what AI can and cannot do; and keep the instructor responsible for instructional decisions.

\subsection{Socio-Technical Systems}

Socio-technical systems theory holds that performance depends on the joint optimization of social and technical systems rather than on technology alone (Mumford, 2000; Trist \& Bamforth, 1951). In the DOT Framework this principle appears in the macro-system level: classroom innovation is expected to be fragile unless policy, training, and infrastructure support it.

\subsection{Using the Framework as an Analytic Lens}

Table 1 shows how each section of the survey corresponds to the constructs above, to elements of the DOT Framework, and to the research questions. Because the instrument was written by the framework's authors to reflect its elements, recovering a framework-consistent structure in the data cannot be treated as independent confirmation of the framework. We therefore use the framework as an interpretive lens and describe correspondences rather than test causal or confirmatory claims.

\begin{table}[!htbp]
\caption{Alignment of Survey Sections With Constructs, DOT Framework Elements, and Research Questions}
\label{tab:alignment}
\small
\begin{tabularx}{\textwidth}{@{}>{\raggedright\arraybackslash}p{0.29\textwidth} >{\raggedright\arraybackslash}p{0.19\textwidth} >{\raggedright\arraybackslash}X >{\raggedright\arraybackslash}p{0.12\textwidth}@{}}
\toprule
Survey section (items) & Construct & DOT Framework element & RQ \\
\midrule
AI familiarity; professional development status (2) & Experience and preparation & Input (micro); Structures (macro) & RQ1, RQ5 \\\addlinespace
Current AI use across 11 tasks (11) & AI use & Process and Output (micro) & RQ1 \\\addlinespace
Design-oriented activities (4) & Design-oriented practice & Empathize, Define, Ideate/Prototype, Test $\times$ Feedback (micro) & RQ2 \\\addlinespace
Prompting techniques (6 options) & Verification and refinement & Prototype and Test $\times$ Process (micro) & RQ2 \\\addlinespace
Training priorities (4) & Support needs & Structures (macro) & RQ2, RQ4 \\\addlinespace
Beliefs about AI's pedagogical role (19) & Practitioner beliefs & Co-intelligence principles; micro and macro levels & RQ3, RQ5 \\\addlinespace
Institutional challenges (6 options); concerns (12); implementation-model preferences (6 options) & Institutional conditions & Environment, Structures, Feedback (macro) & RQ4 \\\addlinespace
Open comment (1) & Illustrative perspectives & Not mapped & Illustrative only \\
\bottomrule
\end{tabularx}
\floatnote{\textit{Note.} DOT = Design Thinking, Open Systems, Transformation. Correspondences reflect the authors' design intent and are interpretive.}
\end{table}

\section{Method}

\subsection{Design}

We used an exploratory, cross-sectional survey design. Descriptive analyses addressed RQ1, RQ2, and RQ4; exploratory factor analysis addressed RQ3; and rank-order correlations addressed RQ5.

\subsection{Participants and Procedure}

Participants were recruited through educational technology conferences, email lists, professional networks, and social media. The survey was administered online through Google Forms between March 10 and June 11, 2025. All 72 respondents read an information form approved by the Institutional Review Board of East Stroudsburg University and consented to participate. Because recruitment relied on educational technology channels, the sample is likely to over-represent practitioners who are engaged with AI.

Most respondents were full-time faculty (\textit{n} = 39, 54.2\%); others were instructional designers (\textit{n} = 10, 13.9\%), adjunct or part-time faculty (\textit{n} = 9, 12.5\%), administrators (\textit{n} = 7, 9.7\%), graduate assistants (\textit{n} = 3, 4.2\%), students (\textit{n} = 3, 4.2\%), and one K--12 teacher (1.4\%). Over three quarters reported actively using AI in teaching (\textit{n} = 37, 51.4\%) or identified as advanced users or researchers (\textit{n} = 18, 25.0\%); eight (11.1\%) were somewhat familiar with AI, seven (9.7\%) were familiar but not using it, and two (2.8\%) were not familiar with AI in education.

\subsection{Instrument}

The DOT Survey was developed by the first two authors to assess beliefs, practices, and institutional conditions related to AI integration, with items written to reflect the design thinking stages, open systems components, and co-intelligence principles of the DOT Framework. No items were adapted from existing instruments. The survey did not ask which AI tools respondents used; it focused on whether reflective, design-oriented, and systems-aware practices were present in their AI use. Before administration, the items were reviewed by 3 collaborating experts before and after a pilot administration took place.

Excluding consent and an optional contact field, the instrument contained 13 questions (Table 1): three background questions (role, AI familiarity, and professional development status); five grid questions comprising 50 rated items (11 usage items, 19 belief items, 4 design-oriented activity items, 4 training-priority items, and 12 concern items); three select-all-that-apply checklists (prompting techniques, institutional challenges, and valued features of an implementation model); one optional open-ended comment item; and one question about interest in a follow-up pilot study, which we do not analyze. The full instrument is available at the link in the Appendix.

Responses were coded as follows. Belief items used a 5-point agreement scale (1 = \textit{strongly disagree}, 2 = \textit{disagree}, 3 = \textit{uncertain}, 4 = \textit{agree}, 5 = \textit{strongly agree}). Usage items used a 5-point scale (1 = \textit{do not use}, 2 = \textit{planning to use}, 3 = \textit{have started using}, 4 = \textit{use occasionally}, 5 = \textit{use frequently}). Design-oriented activities used a 3-point scale (1 = \textit{never}, 2 = \textit{sometimes}, 3 = \textit{always}). Training priorities and concerns were categorical (e.g., \textit{top priority}, \textit{nice to have}, \textit{not important}; \textit{top concern}, \textit{secondary concern}, \textit{not concerned}, \textit{unsure}). Each checklist option was coded 0 or 1. Familiarity was treated as ordinal from 1 (\textit{not familiar at all}) to 5 (\textit{advanced user or researcher}). There were no missing data on rated items. Five, four, and two respondents left the three checklists blank, respectively; to keep denominators consistent, we report all checklist percentages out of the full sample (\textit{N} = 72).

\subsection{Data Analysis}

For RQ1, RQ2, and RQ4 we computed frequencies, percentages, means, and standard deviations. For RQ3, we assessed the suitability of the 19 belief items with the Kaiser-Meyer-Olkin (KMO) measure and Bartlett's test of sphericity and then conducted exploratory factor analysis using principal axis factoring (Fabrigar et al., 1999). Because beliefs about AI were expected to be correlated, we used Promax rotation ($\kappa$ = 4) as the primary solution and Varimax rotation as a robustness check. We determined the number of factors by considering parallel analysis (Horn, 1965) based on 1,000 random data sets and the 95th percentile eigenvalues, the Kaiser criterion, and interpretability. Items with a primary pattern coefficient of at least .40 were assigned to a factor. We estimated internal consistency with Cronbach's alpha. As a sensitivity check, we repeated the analysis without the two items whose individual KMO values fell below .50.

For RQ5, we computed Spearman rank correlations between familiarity and five composite scores: the mean of the items assigned to each belief factor, the mean of the 11 usage items, and the mean of the four design-oriented activity items. We adjusted \textit{p} values for the five tests using the Holm procedure. Responses to the open-ended item were not systematically coded; we quote a small number of anonymized responses, with participants' consent, for illustration only.

\section{Results}

\subsection{RQ1: AI Familiarity and Use}

As described above, 76.4\% of respondents were active or advanced AI users. Regarding professional development, 29 respondents (40.3\%) reported self-directed learning, 23 (31.9\%) had received institution-provided professional development, 15 (20.8\%) had received none but wanted it, and five (6.9\%) had received none and were not interested.

Use was highest for generative and content-related tasks and lowest for evaluative tasks (Table 2). Generating instructional materials had the highest mean (\textit{M} = 3.43, \textit{SD} = 1.43), with 60\% of respondents using AI for this occasionally or frequently. Automating grading had the lowest mean (\textit{M} = 1.60, \textit{SD} = 1.11); 69\% did not use AI for this purpose. About half of respondents (49\%) did not use AI to provide personalized student feedback.

\begin{table}[!htbp]
\caption{Current AI Use by Task Category}
\label{tab:use}
\small
\begin{tabularx}{\textwidth}{@{}X r r c c@{}}
\toprule
Task category & \textit{M} & \textit{SD} & Do not use (\%) & Occasionally or frequently (\%) \\
\midrule
Generating instructional materials & 3.43 & 1.43 & 17 & 60 \\
Supporting own research & 3.21 & 1.43 & 19 & 44 \\
Generating lesson plans & 2.97 & 1.50 & 29 & 47 \\
Developing role-plays, simulations, or case studies & 2.92 & 1.48 & 22 & 43 \\
Supporting student research and academic writing & 2.86 & 1.36 & 22 & 40 \\
Automating administrative tasks & 2.58 & 1.45 & 32 & 29 \\
Assisting in assessments & 2.26 & 1.42 & 47 & 24 \\
Providing personalized student feedback & 2.17 & 1.45 & 49 & 24 \\
Assisting in grading & 1.82 & 1.26 & 61 & 14 \\
Automating assessments & 1.68 & 1.15 & 67 & 11 \\
Automating grading & 1.60 & 1.11 & 69 & 8 \\
\bottomrule
\end{tabularx}
\floatnote{\textit{Note. N} = 72. Scale: 1 = \textit{do not use}, 2 = \textit{planning to use}, 3 = \textit{have started using}, 4 = \textit{use occasionally}, 5 = \textit{use frequently}. Rows are ordered by mean.}
\end{table}

\subsection{RQ2: Design-Oriented Practices and Prompting Techniques}

Experimenting with different prompts was the most frequently reported design-oriented practice, with 70.8\% doing so always (Table 3). Collecting feedback and iterating on AI outputs was the least frequent: half of respondents did so always, and 18.1\% never did. About 60\% always identified needs before using AI, and fewer than half (48.6\%) always outlined what AI should accomplish before implementation.

\begin{table}[!htbp]
\caption{Frequency of Design-Oriented Practices When Using AI}
\label{tab:design}
\small
\begin{tabularx}{\textwidth}{@{}>{\raggedright\arraybackslash}X c c c r r@{}}
\toprule
Practice & Never \textit{n} (\%) & Sometimes \textit{n} (\%) & Always \textit{n} (\%) & \textit{M} & \textit{SD} \\
\midrule
I identify student, faculty, or institutional needs before using AI & 5 (6.9) & 24 (33.3) & 43 (59.7) & 2.53 & 0.63 \\\addlinespace
I clearly outline what AI should accomplish before implementation & 4 (5.6) & 33 (45.8) & 35 (48.6) & 2.43 & 0.60 \\\addlinespace
I experiment with different AI prompts and responses & 4 (5.6) & 17 (23.6) & 51 (70.8) & 2.65 & 0.59 \\\addlinespace
I collect feedback and iterate on AI-generated outputs & 13 (18.1) & 23 (31.9) & 36 (50.0) & 2.32 & 0.77 \\
\bottomrule
\end{tabularx}
\floatnote{\textit{Note. N} = 72. Scale: 1 = \textit{never}, 2 = \textit{sometimes}, 3 = \textit{always}.}
\end{table}

Among prompting techniques, respondents most often reported structuring prompts with progressive refinement (\textit{n} = 59, 81.9\%), cross-checking AI responses with other sources (\textit{n} = 57, 79.2\%), and providing context and constraints (\textit{n} = 57, 79.2\%). Using examples and format expectations (\textit{n} = 49, 68.1\%) and role-based priming (\textit{n} = 44, 61.1\%) were less common. Two respondents (2.8\%) reported accepting AI responses without further verification. Asked about training priorities, respondents most often rated understanding bias and limitations in AI responses as a top priority (\textit{n} = 45, 62.5\%), followed by strategies for iterating on and refining AI-generated content (\textit{n} = 35, 48.6\%), techniques for psychological priming (\textit{n} = 32, 44.4\%), and how to phrase questions (\textit{n} = 24, 33.3\%).

\subsection{RQ3: Structure of Beliefs About AI's Pedagogical Role}

\subsubsection{Factorability and Factor Retention}

Bartlett's test was significant, $\chi^2(171)$ = 886.84, \textit{p} $<$ .001, and the overall KMO was .81. Two items had individual KMO values below .50: institutional support for AI integration (.45) and faculty control over AI-generated materials (.42). The first three eigenvalues were 7.62, 2.50, and 1.35, and five eigenvalues exceeded 1.0. Parallel analysis supported two factors, because the third eigenvalue fell below its 95th percentile random-data value (1.78). We retained three factors for three reasons. First, the third factor formed a coherent cluster of collaboration and planning items that was conceptually distinct from beliefs about task-level capabilities and corresponded to the collaborative element of the DOT Framework. Second, the three-factor structure was stable: when the two low-KMO items were removed, the same three item clusters re-emerged, except that the higher-order thinking item then divided its loading between the first and third factors (.32 and .37). Third, Varimax rotation reproduced the same clusters, with larger cross-loadings, as expected when factors are correlated. Because parallel analysis favored fewer factors, we treat the third factor as provisional.

\subsubsection{Factor Structure}

Table 4 reports the Promax pattern coefficients, communalities, and item descriptive statistics. Together, the three factors accounted for 54.4\% of the item variance. Under Varimax rotation, where factor contributions are additive, they accounted for 24.8\%, 12.5\%, and 17.1\%, respectively. Three pattern coefficients exceeded 1.0 (collaborate with an instructor, 1.065; efficiency in feedback, 1.027; effectiveness of feedback, 1.006). With correlated factors, pattern coefficients are regression weights rather than correlations and can exceed 1.0; these items' communalities were all below 1.0, so they are not improper (Heywood) solutions.

\begin{table}[!htbp]
\caption{Promax Pattern Coefficients, Communalities, and Descriptive Statistics for the 19 Belief Items}
\label{tab:efa}
\small
\setlength{\tabcolsep}{4.5pt}
\begin{tabularx}{\textwidth}{@{}>{\raggedright\hangindent=1em\arraybackslash}X r r r r r r c@{}}
\toprule
Item: ``AI can\ldots'' unless stated & F1 & F2 & F3 & $h^2$ & \textit{M} & \textit{SD} & Agree (\%) \\
\midrule
\multicolumn{8}{@{}l}{\textit{Factor 1: AI Functional Capabilities ($\alpha$ = .915)}} \\
improve efficiency in feedback & \textbf{1.027} & .055 & $-$.243 & .784 & 3.64 & 1.07 & 64 \\
improve effectiveness of feedback & \textbf{1.006} & $-$.078 & $-$.134 & .788 & 3.44 & 1.05 & 53 \\
improve effectiveness of grading & \textbf{.785} & .026 & .071 & .719 & 3.26 & 1.09 & 46 \\
improve efficiency in grading & \textbf{.774} & $-$.014 & .069 & .673 & 3.51 & 0.98 & 54 \\
enhance creativity & \textbf{.586} & .146 & .132 & .565 & 3.72 & 0.97 & 65 \\
serve as a teaching assistant & \textbf{.505} & .043 & .241 & .514 & 3.43 & 1.10 & 60 \\
enhance higher-order thinking & \textbf{.439} & $-$.024 & .273 & .432 & 3.56 & 1.07 & 61 \\
\addlinespace
\multicolumn{8}{@{}l}{\textit{Factor 2: Oversight and Governance ($\alpha$ = .786)}} \\
AI should be integrated with transparency and institutional guidelines & .041 & \textbf{.847} & $-$.098 & .705 & 4.60 & 0.74 & 92 \\
AI-generated content should be reviewed and refined by instructors before use & $-$.187 & \textbf{.781} & .277 & .652 & 4.65 & 0.82 & 92 \\
AI-generated materials need to be critically evaluated by instructors & .336 & \textbf{.761} & $-$.267 & .708 & 4.79 & 0.60 & 97 \\
Faculty should maintain full control over AI-generated instructional materials & $-$.265 & \textbf{.550} & .062 & .268 & 4.36 & 0.94 & 83 \\
\addlinespace
\multicolumn{8}{@{}l}{\textit{Factor 3: Instructor Collaboration and Planning ($\alpha$ = .851)}} \\
collaborate with an instructor & $-$.309 & $-$.098 & \textbf{1.065} & .725 & 4.10 & 0.87 & 82 \\
support instructors & .088 & $-$.183 & \textbf{.711} & .561 & 4.43 & 0.58 & 99 \\
improve effectiveness of instructional planning & .075 & .181 & \textbf{.685} & .656 & 3.68 & 1.03 & 65 \\
improve efficiency in instructional planning & .044 & .210 & \textbf{.666} & .610 & 4.01 & 0.78 & 81 \\
enhance inclusivity in higher education & .302 & $-$.090 & \textbf{.482} & .503 & 3.50 & 1.03 & 46 \\
\addlinespace
\multicolumn{8}{@{}l}{\textit{Items not assigned to a factor}} \\
automate tasks & .379 & .215 & .055 & .286 & 4.18 & 0.86 & 85 \\
replace instructor expertise in some cases & .228 & $-$.193 & .238 & .171 & 2.18 & 1.01 & 14 \\
My institution supports AI integration in teaching and learning & .089 & $-$.080 & .008 & .010 & 3.94 & 0.80 & 71 \\
\bottomrule
\end{tabularx}
\floatnote{\textit{Note. N} = 72. Principal axis factoring with Promax rotation ($\kappa$ = 4). Bold values are primary pattern coefficients of .40 or greater. $h^2$ = communality. Agree = percentage responding \textit{agree} or \textit{strongly agree}. Factor correlations: F1 and F2 = .35, F1 and F3 = .73, F2 and F3 = .26. Alpha for all 19 items = .902.}
\end{table}

Factor 1, \textit{AI Functional Capabilities}, comprised seven items about AI's value for feedback, grading, creativity, teaching assistance, and higher-order thinking ($\alpha$ = .915). Item means were moderate (3.26 to 3.72), and agreement ranged from 46\% to 65\%. Factor 2, \textit{Oversight and Governance}, comprised four items about transparency, institutional guidelines, review, and faculty control ($\alpha$ = .786). These items had the highest means (4.36 to 4.79), and three of them were endorsed by 92\% to 97\% of respondents, indicating ceiling effects. Factor 3, \textit{Instructor Collaboration and Planning}, comprised five items about AI collaborating with and supporting instructors, improving planning, and enhancing inclusivity ($\alpha$ = .851; item means 3.50 to 4.43). Three items were not assigned to a factor. Most respondents agreed that AI can automate tasks (85\%), whereas few agreed that AI can replace instructor expertise in some cases (14\%; 71\% disagreed). The institutional support item had a near-zero communality (.010).

\subsection{RQ4: Institutional Challenges, Concerns, and Support Needs}

The most frequently reported institutional challenges were uncertainty about ethical concerns and academic integrity (63.9\%) and insufficient training or professional development (61.1\%; Table 5). Every listed challenge was selected by more than a third of respondents. Four respondents wrote in additional challenges, three of which concerned training. Ethical, integrity, and critical-thinking risks were the most frequently rated top concerns, whereas workload, resistance, and alignment with learning goals were rated as top concerns least often.

\begin{table}[!htbp]
\caption{Institutional Challenges Encountered and Concerns Rated as Top Concerns}
\label{tab:challenges}
\small
\begin{tabularx}{\textwidth}{@{}X r r@{}}
\toprule
Item & \textit{n} & \% \\
\midrule
\multicolumn{3}{@{}l}{\textit{Institutional challenges encountered (select all that apply)}} \\
Uncertainty about ethical concerns and academic integrity & 46 & 63.9 \\
Insufficient training or professional development & 44 & 61.1 \\
Lack of institutional policies on AI use & 36 & 50.0 \\
Limited access to AI tools and infrastructure & 34 & 47.2 \\
Resistance from faculty or administrators & 32 & 44.4 \\
Lack of alignment between AI tools and institutional priorities & 27 & 37.5 \\
\addlinespace
\multicolumn{3}{@{}l}{\textit{Concerns rated as a top concern}} \\
Concerns about bias, ethics, or academic integrity & 50 & 69.4 \\
Ethical concerns (e.g., data privacy, academic dishonesty) & 45 & 62.5 \\
Over-reliance on AI reducing student critical thinking & 42 & 58.3 \\
AI-generated bias in content & 34 & 47.2 \\
Equity in access to AI tools across institutions & 33 & 45.8 \\
Lack of clarity on how to use AI effectively & 25 & 34.7 \\
Lack of professional development or training & 25 & 34.7 \\
Limited institutional policies or guidelines & 19 & 26.4 \\
Uncertainty about what AI tools are available and appropriate & 18 & 25.0 \\
AI's impact on faculty workload and role changes & 17 & 23.6 \\
Resistance from faculty, administrators, or students & 14 & 19.4 \\
Difficulty in aligning AI with educational goals and learning outcomes & 13 & 18.1 \\
\bottomrule
\end{tabularx}
\floatnote{\textit{Note.} Percentages are based on the full sample (\textit{N} = 72). Four respondents left the challenges checklist blank.}
\end{table}

When asked which features of an AI implementation model would be most valuable, respondents most often selected examples of best practices from higher education institutions (\textit{n} = 55, 76.4\%) and ethical considerations and risk management strategies (\textit{n} = 52, 72.2\%), followed by feedback loops for continuous improvement and training on prompt engineering and customization (each \textit{n} = 46, 63.9\%), institutional strategies for adoption at scale (\textit{n} = 39, 54.2\%), and step-by-step integration guidance (\textit{n} = 37, 51.4\%). Responses to the optional open-ended item were few, but two illustrate the gap some respondents described between their own practice and their institutional context: ``I am ahead of the curve compared to administration and other faculty so that any support is too basic,'' and ``I'm struggling with getting colleagues to even try to use AI. It's like they are terrified.''

\subsection{RQ5: Differences by AI Familiarity}

AI familiarity was positively associated with AI use ($r_s$ = .57, \textit{p} $<$ .001), design-oriented practice ($r_s$ = .37, \textit{p} = .001), Functional Capabilities beliefs ($r_s$ = .33, \textit{p} = .005), and Collaboration and Planning beliefs ($r_s$ = .29, \textit{p} = .015). All four associations remained significant after Holm adjustment (adjusted \textit{p}s $\le$ .030). Familiarity was not associated with Oversight and Governance beliefs ($r_s$ = .06, \textit{p} = .613). The seven respondents who were familiar with AI but not using it had the lowest means of any familiarity group for AI use (\textit{M} = 1.39), design-oriented practice (\textit{M} = 1.96), Functional Capabilities (\textit{M} = 2.63), and Collaboration and Planning (\textit{M} = 3.23), while their Oversight and Governance mean remained high (\textit{M} = 4.71). Because group sizes ranged from 2 to 37, we report these group means descriptively.

\section{Discussion}

We discuss the findings by research question and then consider how the belief structure corresponds to the DOT Framework. Throughout, we distinguish what the data show from our interpretation of them.

\subsection{Use Concentrated in Generative Tasks (RQ1)}

Respondents used AI mainly for generating materials and supporting research and rarely for grading, assessment, or personalized feedback. This pattern is consistent with faculty surveys that emphasize content creation, research support, and administrative efficiency (Almisad \& Aleidan, 2025). One interpretation, consistent with the strong endorsement of oversight items, is that practitioners are reluctant to delegate evaluative judgments; however, the survey did not ask why respondents avoided evaluative uses, and limited access to suitable tools is an equally plausible explanation.

\subsection{An Incomplete Design Cycle (RQ2)}

In DOT terms, respondents reported the Ideate and Prototype stages (experimenting with prompts and refining outputs) more consistently than the Test $\times$ Feedback intersection (collecting feedback and iterating), and only about half always specified goals before using AI. We interpret this as evidence that much current AI use is exploratory and locally refined rather than designed and evaluated as a cycle. Verification was nevertheless widespread at the level of individual outputs: most respondents cross-checked AI responses, and almost none accepted responses without verification. The gap therefore seems to lie less in checking individual outputs than in gathering evidence from students and colleagues about whether AI-supported designs work. The high priority respondents gave to training on bias and limitations is consistent with the call to guide responsible use and verification (Melisa et al., 2025).

\subsection{Belief Structure and the DOT Framework (RQ3)}

Table 6 summarizes how we interpret each factor in relation to the framework. These correspondences are interpretive: the items were written from the framework, so their clustering is consistent with its distinctions but does not independently confirm them.

\begin{table}[!htbp]
\caption{Interpretive Correspondence Between Belief Factors and DOT Framework Elements}
\label{tab:correspondence}
\small
\begin{tabularx}{\textwidth}{@{}>{\raggedright\arraybackslash}p{0.17\textwidth} >{\raggedright\arraybackslash}X >{\raggedright\arraybackslash}X >{\raggedright\arraybackslash}p{0.24\textwidth}@{}}
\toprule
Factor & Content of items & Corresponding DOT element & Co-intelligence principle \\
\midrule
F1: AI Functional Capabilities & What AI can do for feedback, grading, creativity, teaching assistance, and higher-order thinking & Process and Output at the Prototype and Test stages; micro level & AI as a capable but fallible collaborator; support rather than replace higher-order thinking \\\addlinespace
F2: Oversight and Governance & Review and critical evaluation of AI outputs; transparency, institutional guidelines, and faculty control & Test $\times$ Feedback and Structures; spans micro (personal review) and macro (institutional guidelines) levels & Human responsibility for instructional decisions \\\addlinespace
F3: Instructor Collaboration and Planning & AI collaborating with and supporting instructors, improving planning, and enhancing inclusivity & Input $\times$ Process at the Define and Ideate stages; inclusivity relates to Empathize $\times$ Environment & AI as collaborator; active experimentation \\
\bottomrule
\end{tabularx}
\floatnote{\textit{Note.} Correspondences reflect the authors' interpretation and were not tested statistically.}
\end{table}

Three observations follow. First, Oversight and Governance emerged as a distinct and near-universally endorsed dimension that did not vary with familiarity. We read this as a shared professional norm rather than a belief that develops with experience, although ceiling effects and social desirability may also contribute. Its lower reliability may reflect the fact that its items combine personal review practices with institutional expectations, two levels the framework separates. Second, the planning items clustered with the collaboration items rather than with the task-capability items, which suggests that respondents may regard planning with AI as a joint activity; this interpretation is tentative. Third, the strong correlation between Factors 1 and 3 (.73), together with the parallel analysis result, means that these two factors may not be empirically distinct in other samples. Discriminant validity is therefore the most important question for confirmatory work. Finally, the two unassigned items behaved differently from the others: the replacement item was rejected by most respondents and may function as a boundary condition for co-intelligence rather than as a belief that varies, and the institutional support item may describe a context rather than a belief about AI's pedagogical role.

\subsection{Institutional Conditions (RQ4)}

The challenges respondents reported were spread across ethics, training, policy, infrastructure, and resistance, and every listed challenge was reported by more than a third of the sample. In the framework's terms, these are macro-system conditions. The pattern is consistent with the socio-technical expectation that classroom innovation depends on supporting structures (Mumford, 2000) and with other reports of insufficient institutional guidance (Baytas \& Ruediger, 2025). Respondents' preference for examples of best practice and for ethical guidance over step-by-step instructions suggests that they want contextualized models rather than procedures, but the survey cannot show which supports would change practice.

\subsection{The Role of Familiarity (RQ5)}

Familiarity was associated with use, design-oriented practice, and capability and collaboration beliefs, but not with oversight beliefs. Because the design is cross-sectional, these associations may reflect experience shaping beliefs, favorable beliefs leading to experience, or both. The low scores of respondents who were familiar with AI but not using it suggest that awareness without use may not be enough to develop the capability beliefs or design practices described above. This finding is based on seven respondents and needs replication.

\section{Limitations}

The study has several limitations. First, the sample was small (\textit{N} = 72) and self-selected through educational technology channels, and most respondents were active AI users; the findings describe engaged practitioners and should not be generalized to the wider workforce. Some subgroups, such as K--12 educators and administrators, were very small. Second, all data were self-reported and may overstate principled practice, particularly for oversight and verification items. Third, the design was cross-sectional and cannot establish causal relationships or capture how practices change over time. Fourth, the instrument was developed for this study, was not adapted from validated measures, and was written by the framework's authors; the factor structure is therefore exploratory, and the correspondence with the framework is not independent evidence for it. A sample of 72 is small for factor analysis, parallel analysis favored two factors, the F1 and F3 factors were highly correlated, two items had low sampling adequacy, the oversight items showed ceiling effects, and the analysis treated 5-point ordinal items as continuous. The instrument also does not clearly separate micro-level practices from macro-level conditions, and it contains no reverse-coded items. Fifth, the open-ended responses were few and were not systematically analyzed. Confirmatory factor analysis in larger, independent, and more diverse samples is needed to test the factor structure, including the discriminant validity of the first and third factors.

\section{Implications}

\subsection{For Practitioners}

The weakest part of the design cycle was the Test $\times$ Feedback intersection. Practitioners who already experiment with prompts and verify outputs could add two steps at either end of that cycle: stating what AI is meant to accomplish before using it, and collecting evidence from students or colleagues afterward. Because many respondents learned on their own, peer formats in which colleagues share prompts, outputs, and evidence of effectiveness may be a practical way to add these steps.

\subsection{For Institutions}

Institutional challenges were widespread, and respondents most valued examples of best practice and ethical guidance. Institutions could prioritize clear usage guidelines, curated examples from comparable institutions, and shared criteria for evaluating AI-supported teaching. The association between familiarity and practice suggests that professional development may need to differ for practitioners who are aware of AI but not using it and for experienced users.

\subsection{For Researchers}

Three directions follow from this study. First, the belief structure should be tested with confirmatory factor analysis in independent samples, with items revised to separate micro-level review practices from macro-level governance and to reduce ceiling effects. Second, studies should link beliefs and design practices to the quality of AI-supported instructional products, for example through rubric-based evaluation of artifacts or analysis of instructor-AI interactions. Third, design-based research could test whether structured supports aligned with the framework change practice compared with unstructured AI use.

\section{Conclusion}

In this sample of engaged higher education practitioners, AI was used mainly for generative tasks rather than evaluative ones (RQ1). Experimentation was common, but specifying goals in advance and collecting feedback on outcomes were less consistent, which indicates an incomplete design cycle (RQ2). Beliefs about AI's pedagogical role formed three provisional dimensions (AI Functional Capabilities, Oversight and Governance, and Instructor Collaboration and Planning) that are consistent with, but do not confirm, the distinctions of the DOT Framework (RQ3). Practitioners reported widespread institutional challenges, especially regarding ethics, training, and policy (RQ4), and AI familiarity was associated with use, design practice, and capability beliefs but not with oversight beliefs (RQ5). Together, these findings suggest where the framework may be most useful in practice, in strengthening the needs-definition and feedback ends of the design cycle and its institutional supports, and they identify the measurement questions that confirmatory research should address next.

\section{References}
\begingroup
\RaggedRight
\begin{list}{}{\setlength{\leftmargin}{0.5in}\setlength{\itemindent}{-0.5in}\setlength{\itemsep}{2pt}\setlength{\parsep}{0pt}\setlength{\topsep}{0pt}}

\item Almisad, B., \& Aleidan, A. (2025). Faculty perspectives on generative artificial intelligence: Insights into awareness, benefits, concerns, and uses. \textit{Frontiers in Education, 10}, Article 1632742. \url{https://doi.org/10.3389/feduc.2025.1632742}

\item Azukas, E., \& Gibson, D. (2025). Co-intelligence in the classroom: The DOT Framework for AI-enhanced teaching and learning. \textit{AI Enhanced Learning, 1}(2), 269--293. \url{https://doi.org/10.70725/846858waolmj}

\item Batista, J., Mesquita, A., \& Carnaz, G. (2024). Generative AI and higher education: Trends, challenges, and future directions from a systematic literature review. \textit{Information, 15}(11), Article 676. \url{https://doi.org/10.3390/info15110676}

\item Baytas, C., \& Ruediger, D. (2024). \textit{Generative AI in higher education: The product landscape}. Ithaka S+R. \url{https://doi.org/10.18665/sr.320394}

\item Baytas, C., \& Ruediger, D. (2025). \textit{Making AI generative for higher education: Adoption and challenges among instructors and researchers}. Ithaka S+R. \url{https://doi.org/10.18665/sr.322677}

\item Bond, M., Khosravi, H., De Laat, M., Bergdahl, N., Negrea, V., Oxley, E., Pham, P., Chong, S. W., \& Siemens, G. (2024). A meta systematic review of artificial intelligence in higher education: A call for increased ethics, collaboration, and rigour. \textit{International Journal of Educational Technology in Higher Education, 21}, Article 4. \url{https://doi.org/10.1186/s41239-023-00436-z}

\item Davis, F. D. (1989). Perceived usefulness, perceived ease of use, and user acceptance of information technology. \textit{MIS Quarterly, 13}(3), 319--340. \url{https://doi.org/10.2307/249008}

\item Ertmer, P. A. (2005). Teacher pedagogical beliefs: The final frontier in our quest for technology integration? \textit{Educational Technology Research and Development, 53}(4), 25--39. \url{https://doi.org/10.1007/BF02504683}

\item Fabrigar, L. R., Wegener, D. T., MacCallum, R. C., \& Strahan, E. J. (1999). Evaluating the use of exploratory factor analysis in psychological research. \textit{Psychological Methods, 4}(3), 272--299. \url{https://doi.org/10.1037/1082-989X.4.3.272}

\item Gibson, D. C., Azukas, M. E., Yilmaz Soylu, M., \& Lee, J. (2026). Perspective chapter: Leading organizational transformation in the age of artificial intelligence: The DOT Framework for co-intelligent leadership. In \textit{Studies of leadership in the digital age}. IntechOpen. \url{https://doi.org/10.5772/intechopen.1016262}

\item Henriksen, D., Richardson, C., \& Mehta, R. (2017). Design thinking: A creative approach to educational problems of practice. \textit{Thinking Skills and Creativity, 26}, 140--153. \url{https://doi.org/10.1016/j.tsc.2017.10.001}

\item Horn, J. L. (1965). A rationale and test for the number of factors in factor analysis. \textit{Psychometrika, 30}(2), 179--185. \url{https://doi.org/10.1007/BF02289447}

\item Luhmann, N. (1995). \textit{Social systems} (J. Bednarz, Jr., with D. Baecker, Trans.). Stanford University Press. \url{https://www.sup.org/books/title?id=2225} (Original work published 1984)

\item Lyu, W., Zhang, S., Chung, T., Sun, Y., \& Zhang, Y. (2025). Understanding the practices, perceptions, and (dis)trust of generative AI among instructors: A mixed-methods study in the U.S. higher education. \textit{Computers and Education: Artificial Intelligence, 8}, Article 100383. \url{https://doi.org/10.1016/j.caeai.2025.100383}

\item Mah, D.-K., \& Groß, N. (2024). Artificial intelligence in higher education: Exploring faculty use, self-efficacy, distinct profiles, and professional development needs. \textit{International Journal of Educational Technology in Higher Education, 21}, Article 58. \url{https://doi.org/10.1186/s41239-024-00490-1}

\item Meinel, C., Leifer, L., \& Plattner, H. (Eds.). (2011). \textit{Design thinking: Understand -- improve -- apply}. Springer. \url{https://doi.org/10.1007/978-3-642-13757-0}

\item Melisa, R., Ashadi, A., Triastuti, A., Hidayati, S., Salido, A., Ero, P. E. L., Marlini, C., Zefrin, Z., \& Al Fuad, Z. (2025). Critical thinking in the age of AI: A systematic review of AI's effects on higher education. \textit{Educational Process: International Journal, 14}, Article e2025031. \url{https://doi.org/10.22521/edupij.2025.14.31}

\item Mollick, E. (2024). \textit{Co-intelligence: Living and working with AI}. Portfolio. \url{https://www.penguinrandomhouse.com/books/741805/co-intelligence-by-ethan-mollick/9780593716717/}

\item Mumford, E. (2000). A socio-technical approach to systems design. \textit{Requirements Engineering, 5}(2), 125--133. \url{https://doi.org/10.1007/PL00010345}

\item Nevárez Montes, J., \& Elizondo-Garcia, J. (2025). Faculty acceptance and use of generative artificial intelligence in their practice. \textit{Frontiers in Education, 10}, Article 1427450. \url{https://doi.org/10.3389/feduc.2025.1427450}

\item Pea, R. D. (1993). Practices of distributed intelligence and designs for education. In G. Salomon (Ed.), \textit{Distributed cognitions: Psychological and educational considerations} (pp. 47--87). Cambridge University Press. \url{https://web.stanford.edu/~roypea/RoyPDF\%20folder/A67_Pea_93_DI_CUP.pdf}

\item Trist, E. L., \& Bamforth, K. W. (1951). Some social and psychological consequences of the longwall method of coal-getting. \textit{Human Relations, 4}(1), 3--38. \url{https://doi.org/10.1177/001872675100400101}

\item von Bertalanffy, L. (1950). An outline of general system theory. \textit{The British Journal for the Philosophy of Science, 1}(2), 134--165. \url{https://doi.org/10.1093/bjps/I.2.134}

\item Wang, Y., Liu, C., \& Tu, Y.-F. (2021). Factors affecting the adoption of AI-based applications in higher education: An analysis of teachers' perspectives using structural equation modeling. \textit{Educational Technology \& Society, 24}(3), 116--129. \url{https://doi.org/10.30191/ETS.202107_24(3).0009}

\end{list}
\endgroup

\clearpage
\appendix

\section{Appendix}

\subsection{DOT Survey Instrument}

The DOT Survey was developed by M. Elizabeth Azukas (then at East Stroudsburg University, USA) and David Gibson (Curtin University, Australia) under Institutional Review Board oversight and administered through Google Forms. It contains 13 questions covering background and AI familiarity, current AI use (11 tasks), beliefs about AI's pedagogical role (19 items), design-oriented activities (4 items), prompting techniques, training priorities (4 items), institutional challenges, concerns (12 items), valued features of an AI implementation model, an optional open-ended comment, and interest in a follow-up pilot study. The full instrument is available at \url{https://forms.gle/Ax3nWUkG1Lbc3eK58}

\end{document}